\documentclass[letterpaper, 10 pt, conference]{ieeeconf}  
\IEEEoverridecommandlockouts              
\usepackage{booktabs} 
\usepackage{algorithmic}
\usepackage{algorithm}
\usepackage{amssymb}
\usepackage{bm} 
\usepackage{graphicx}
\usepackage{cite}
\usepackage[utf8]{inputenc}
\usepackage[T1]{fontenc}
\usepackage[switch]{lineno}
\usepackage[dvipsnames]{xcolor}
\definecolor{dark_blue}{HTML}{0000A0}
\definecolor{dark_red}{HTML}{8b0000}

\usepackage[colorlinks,
            linkcolor=black,
            anchorcolor=black,
            citecolor=black,
            ]{hyperref}

\usepackage{amsmath}
\DeclareMathOperator*{\argmax}{argmax}

\newcommand{\etal}{\textit{et al. }}

\title{\LARGE \bf
How does the structure embedded in learning policy affect learning quadruped locomotion?
}
\author{Kuangen Zhang$^{1, 2, 3}$, Jongwoo Lee$^{2}$, Zhimin Hou$^{4}$, Clarence W. de Silva$^{3}$, Chenglong Fu$^{1}$, Neville Hogan$^{2}$
\thanks{This work was supported by the National Key R\&D Program of China [Grant 2018YFB1305400]; National Natural Science Foundation of China [Grant U1913205, 61533004 and U1613206]; Guangdong Innovative and Entrepreneurial Research Team Program [Grant 2016ZT06G587]; Shenzhen and Hong Kong Innovation Circle Project [Grant SGLH20180619172011638]; and Centers for Mechanical Engineering Research and Education at MIT and SUSTech.}
\thanks{$^{1}$The Department of Mechanical and Energy Engineering, Southern University of Science and Technology, Shenzhen 518055, China (Corresponding author: Chenglong Fu: fucl@sustech.edu.cn).}%
\thanks{$^{2}$The Department of Mechanical Engineering, Massachusetts Institute of Technology, Cambridge, MA, USA.}%
\thanks{$^{3}$The Department of Mechanical Engineering, University of British Columbia, Vancouver V6T1Z4, Canada.}%
\thanks{$^{4}$The Department of Biomedical Engineering, National University of Singapore, Singapore.}%
\thanks{Code: \url{https://github.com/KuangenZhang/StructuredRL}}
\thanks{Video: \url{https://youtu.be/VmgcgLO2SxQ}
}}

\begin{document}
\maketitle
\thispagestyle{plain}
\pagestyle{plain}

\begin{abstract}
\textcolor{black}{
Reinforcement learning (RL) is a popular data-driven method that has demonstrated great success in robotics. Previous works usually focus on learning an end-to-end (direct) policy to directly output joint torques. While the direct policy seems convenient, the resultant performance may not meet our expectations. To improve its performance, more sophisticated reward functions or more structured policies
can be utilized. This paper focuses on the latter because the structured policy is more intuitive and can inherit insights from previous model-based controllers.
It is unsurprising that the structure, such as a better choice of the action space and constraints of motion trajectory, may benefit the training process and the final performance of the policy at the cost of generality, but the \textit{quantitative effect} is still unclear.
To analyze the effect of the structure quantitatively, this paper investigates three policies with different levels of structure in learning quadruped locomotion: a direct policy, a structured policy, and a highly structured policy.
The structured policy is trained to learn a task-space impedance controller and the highly structured policy learns a controller tailored for trot running, which we adopt from previous work. To evaluate trained policies, we design a simulation experiment to track different desired velocities under force disturbances. 
Simulation results show that structured policy and highly structured policy require 1/3 and 3/4 fewer training steps than the direct policy to achieve a similar level of cumulative reward, and seem more robust and efficient than the direct policy.
We highlight that the structure embedded in the policies significantly affects the overall performance of learning a complicated task when complex dynamics are involved, such as legged locomotion. 
}
\end{abstract}

\section{INTRODUCTION}
\label{sec:introduction}

\textcolor{black}{
Controlling a legged robot is challenging because it is a high dimensional and under-actuated system. To realize the locomotion control, some existing research utilized model-based methods, which included a feedforward trajectory planner and a feedback sensory controller \cite{hyun_high_2014, fu_gait_2008, neunert_whole-body_2018, magana_fast_2019, fahmi_passive_2019}. The model-based method may control the robot to perform well in 
a specific task, but at the cost of generality. Therefore, reinforcement learning (RL) becomes popular in robotics recently because it allows for training a neural network policy to control a robot by learning from its own experience~\cite{lillicrap_continuous_2015, xu_feedback_2019, chen_deep_2019}. 
}
Using RL, mono-, bi-, and quadruped robots
were
successfully trained to walk in rich environments \cite{lillicrap_continuous_2015, zhang_teach_2019}. Combining imitation learning with RL, Peng \etal trained a biped robot in simulation to walk in complex environments and even mimic complex human behaviors, such as backflip and dancing~\cite{peng_deeploco_2017, peng_deepmimic:_2018}. Xie \etal utilized a reference trajectory to train and control a real biped robot (Cassie) to walk in a stable manner~\cite{xie_iterative_2019}. Transferring a neural network policy learned in a simulation environment to the real world has also been successful~\cite{hwangbo_learning_2019, singla_realizing_2019}. 

One of the popular methods in RL for robotics is to learn a direct neural network policy that maps from robot states to joint torques~\cite{haarnoja_soft_2018, fujimoto_addressing_2018, hou_off-policy_2020}. 
\textcolor{black}{
Some researchers also designed a hierarchical framework that consists of a high-level neural network and a low-level neural network to realize locomotion control \cite{jain_hierarchical_2019, hou_off-policy_2020}. 
Because the hierarchical framework does not incorporate any kinematic nor dynamic information of the robot, it can be still regarded as a direct policy.
}
There are several advantages to the direct policy. First, it requires little information of the robot model, hence it can be used for a general class of robots and tasks~\cite{varin_comparison_2019}. Second, in principle, the direct policy does not have any constraint and may potentially outperform constrained policies. 

\textcolor{black}{
On the other hand, the performance of the direct policy may not meet the expectations of researchers. This problem can be solved by adding additional reward functions \cite{tsounis_deepgait_2020, zhang_teach_2019, hwangbo_learning_2019}, but it is hard to find suitable reward functions and tune the weights between different reward functions. Besides, it usually takes a long time to train the direct policy, especially when the robot has a large state-action space. It is often not very intuitive to limit the search space of the direct policy.
}

In order to address these drawbacks, it has been proposed to explore different action spaces by embedding some structures in controller and training policies to learn parameters and/or variables of the controller. Herein \textit{structured policy} denotes this approach. A common example is to implement a joint impedance controller. Peng \etal trained a policy to calculate the target angles of joint impedance controllers to imitate a wide variety of captured motion clips~\cite{peng_deepmimic:_2018}. 
Buchli \etal proposed a PI$^2$ algorithm to learn joint impedance parameters and joint trajectories simultaneously~\cite{buchli_learning_2011}. Bogdanovic \etal demonstrated that
a
variable joint impedance controller for contact-sensitive tasks could outperform
a
direct joint torque controller~\cite{bogdanovic_learning_2019}.

Another alternative is the task-space impedance control. Varin \etal~\cite{varin_comparison_2019} and Martín-Martín \etal~\cite{martin-martin_variable_2019} both demonstrated that the policy to learn task-space impedance control significantly improved sampling efficiency and quality of
a manipulation task compared to joint-space impedance control. The authors reasoned that the former could utilize the robot kinematics and dynamics that are well understood; moreover, working with the task-space controller allowed engineers to intuitively limit the search space by considering the desired workspace of a manipulator. These results show that employing a proper structure heavily affects the sampling efficiency and even the performance of trained behavior. It remains a question whether the same argument can be applied to legged robots. 
\textcolor{black}{
If so, what is the quantitative effect of the structure and how complicated should the structure be.
}

\begin{figure*}[htpb]
    \centering
    \includegraphics[width=0.99\textwidth]{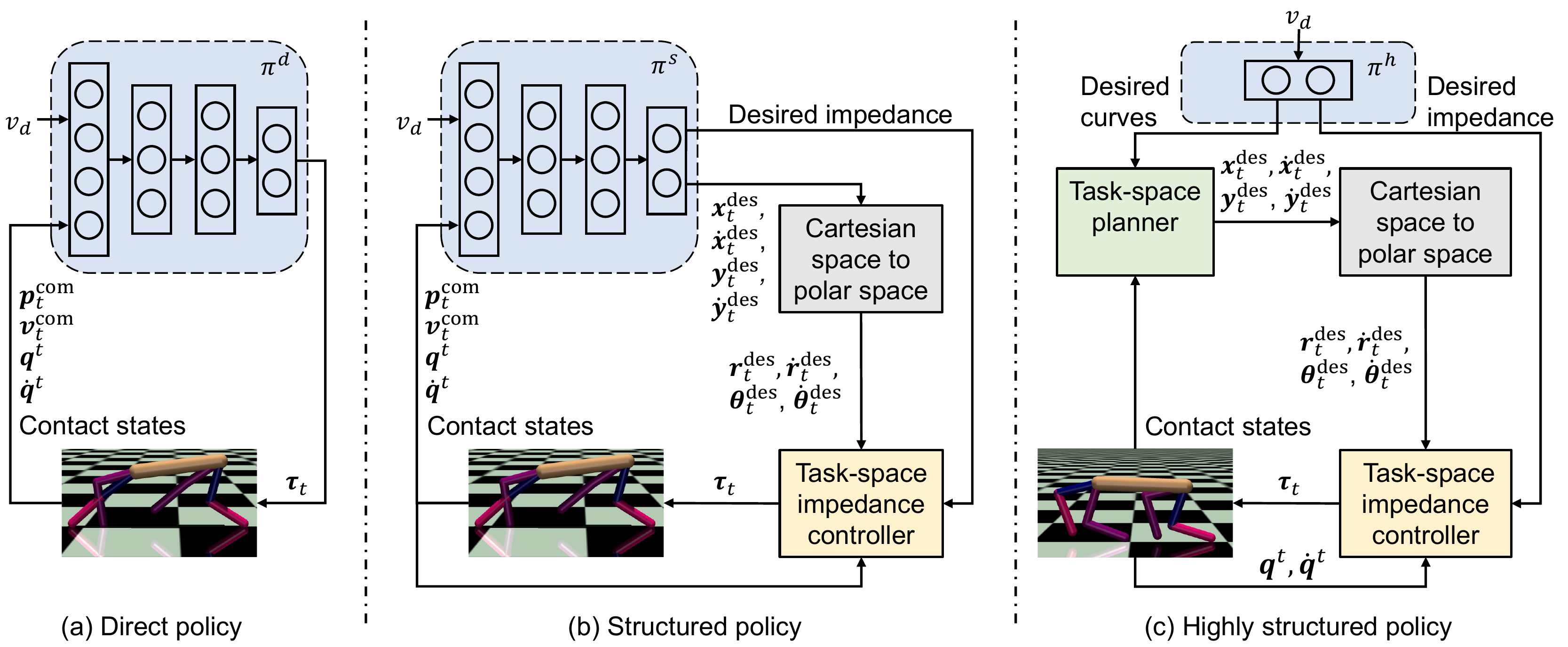}
    \caption{Overview of three different policies.
    } 
    \label{fig:overview} 
\end{figure*}

\textcolor{black}{
In this regard, the present paper focuses on improving the performance of the direct policy by embedding structures and analyzing the quantitative effect of different levels of structure on learning quadruped locomotion. 
This paper compares policies with three different levels of structure: a direct policy, a structured policy, and a highly-structured policy~(\autoref{fig:overview}).
}
The direct policy outputs joint torques as a function of system states and desired velocity without imposing any structure. The structured policy outputs trajectories and impedance parameters of a task-space impedance controller. The highly-structured policy outputs a set of parameters determining task-space impedance and trajectory-scaling parameters of a quadruped locomotion controller that we adopt from previous research~\cite{hyun_high_2014, lee_dynamics_2014}. As 
shown in~\autoref{fig:overview}, the highly-structured policy exploits additional structures tailored to achieve quadrupedal locomotion. 
\textcolor{black}{
The simulation results show that the structured policy and the highly structured policy are able to 1) require 1/3 and 3/4 fewer training steps, 2) increase robustness, and 3) even decrease energetic cost in tracking different desired velocities under disturbances, compared to the direct policy. The highly structured policy also decreases the sensitivity of the policy to the cost function, which is better than the other policies, but its generalization ability is worse than the other policies.
}

\begin{figure*}[htpb]
    \centering
    \includegraphics[width=0.99\textwidth]{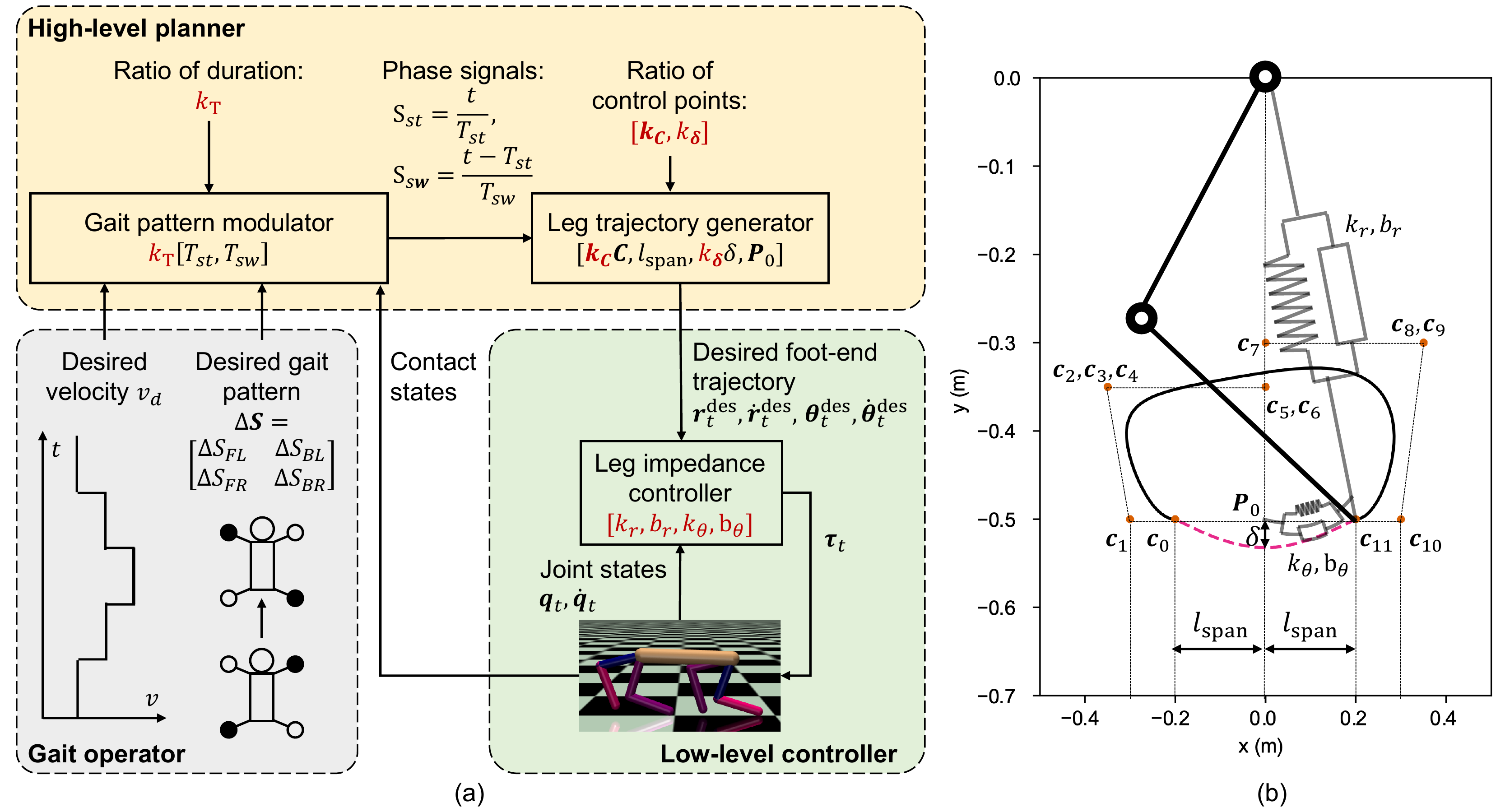}
    \caption{(a) Framework of the task-space planner and controller. 
    Optimized control parameters (see \autoref{tab:parameters}) are marked with red color. 
    (b) The virtual leg impedance in the polar task space and the desired foot-end trajectory. The swing-phase trajectory (black solid line) is a B\'ezier curve controlled by 12 points $\bm{C}$: $\bm{c}_0,\dots, \bm{c}_{11}$. The stance-phase trajectory (red dash line) is a sinusoidal wave whose amplitude and half of the stroke length are $\bm{P}_0$ and $\delta$.} 
    \label{fig:controller} 
\end{figure*}

\begin{table*}[ht]
\centering
\caption {\label{tab:parameters} The definition of control parameters.}
\renewcommand{\arraystretch}{1} 
\begin{center}
\resizebox{1\textwidth}{!}{
\begin{tabular}{l l l l}
\toprule
Terminology & Definition &Terminology & Definition\\
\midrule
$T_\text{st}$ & Desired stance-phase period (s) &
$T_\text{sw}$ 
& Desired swing-phase period (s)\\
$\bm{C}$ & Bézier control points in the swing phase (m) &
$l_\text{span}$ &
Half of the stroke length (m)\\
$\delta$ & Amplitude of the sinusoidal wave in the stance phase (m) &
$P_0$ &
Center point of the stance-phase trajectory (m)\\
$k_r$ &
Radial stiffness of each leg (N/m)&
$b_r$ &
Radial damping of each leg (Ns/m)\\
$k_\theta$ &
Angular stiffness of each leg (Nm/rad)&
$b_\theta$ &
Angular damping of each leg (Nms/rad)\\
\end{tabular}
}
\end{center}
\end{table*}

\section{Policies}
\label{sec:policy}

\subsection{Direct policy}
The direct policy is a common policy in RL \cite{haarnoja_soft_2018, fujimoto_addressing_2018}, which directly maps the states of a robot to the joint torques $\bm{\tau}_t$ (\autoref{fig:overview}a). In our application, the resulting control law is:
\begin{equation}
\label{eq:direct_policy}
\begin{split}
    &\bm{a}^d_t = \pi^d( \bm{s}^d_t),\\
    &\bm{\tau}_t = \bm{a}^d_t, \\
    &\bm{s}^d_t = [v_d, \bm{p}^\text{com}_t, \bm{v}^\text{com}_t, \bm{q}_t, \bm{\dot{q}}_t, \text{contact states}]
\end{split}
\end{equation}
where $\pi_d$ denotes the actor network of the direct policy. $\bm{a}^d_t$ is the action of the actor network. The state $\bm{s}^d_t$ includes the desired velocity $v_d$, center-of-mass height and pitch orientation ($\bm{p}^\text{com}_t$), horizontal and vertical velocity of center of mass and rate of pitch ($\bm{v}^\text{com}_t$), joint angles $\bm{q}_t$ and angular velocities $\bm{\dot{q}}_t$, and contact states of feet (e.g., touchdown and swing). 

\subsection{Structured policy}
The structured policy learns impedance parameters in polar coordinates $(k_r, b_r, k_\theta, b_\theta)$ and desired foot positions ($\bm{x}^\text{des}_{t}, \bm{y}^\text{des}_{t}$) and velocities ($\dot{\bm{x}}^\text{des}_{t}, \dot{\bm{y}}^\text{des}_{t}$) for each leg in Cartesian space~(\autoref{fig:overview}b).

\begin{equation}
    \begin{split}
        &\bm{a}^s_t = \pi^s( \bm{s}^s_t),\\
        &[\bm{x}^\text{des}_{t}, \bm{y}^\text{des}_{t}, 
        \dot{\bm{x}}^\text{des}_{t},
        \dot{\bm{y}}^\text{des}_{t},
        k_r, b_r, k_\theta, b_\theta] = \bm{a}^s_t,\\
        &\bm{s}^s_t = [v_d, \bm{p}^\text{com}_t, \bm{v}^\text{com}_t, \bm{q}_t, \bm{\dot{q}}_t, \text{contact states}],
    \end{split}
\end{equation}
where $\pi^s$ and $\bm{a}^s_t$ denote the actor network and the action of the structured policy. The state $\bm{s}^s_t$ is the same as $\bm{s}^d_t$ in \eqref{eq:direct_policy}.

The task-space impedance controller is defined in polar coordinates
as shown in \autoref{fig:controller}b and the reasoning is described in detail in~\cite{hyun_high_2014, lee_dynamics_2014}. The desired position and velocity of the foot in Cartesian coordinates, obtained from the neural network policy, are converted to polar coordinate variables $(\bm{r}^\text{des}_t, \bm{\theta}^\text{des}_t, \dot{\bm{r}}^\text{des}_t, \dot{\bm{\theta}}^\text{des}_t)$ then
substituted
into the impedance controller. 

\begin{equation}
\label{eq:polar_impedance}
\begin{split}
    &\bm{\tau}_{t,i} = \bm{J}_\text{polar}(\bm{q}_{t,i})^T 
    \begin{bmatrix}
    &k_r (r^\text{des}_{t,i} - r_{t,i})+ b_r (\dot{r}^\text{des}_{t,i} - \dot{r}_{t,i})\\
    &k_\theta (\theta^\text{des}_{t,i} - \theta_{t,i})+ b_\theta (\dot{\theta}^\text{des}_{t,i} - \dot{\theta}_{t,i})\\
    \end{bmatrix},\\
    &[r^\text{des}_{t,i}, \theta^\text{des}_{t,i}, \dot{r}^\text{des}_{t,i}, \dot{\theta}^\text{des}_{t,i}] = 
    T_C^P(x^\text{des}_{t,i}, y^\text{des}_{t,i},\dot{x}^\text{des}_{t,i}, \dot{y}^\text{des}_{t,i}),
\end{split}
\end{equation}
where $\bm{J}_\text{polar}(\bm{q}_{t, i})$ indicates Jacobian matrix of the leg $i$ ($i\in$ \{FL, FR, BL, BR\}). FL, FR, BL, and BR denote front left, front right, back left, and back right, respectively.  $r_{t,i}, \theta_{t,i}, \dot{r}_{t,i}, \dot{\theta}_{t,i}$ indicate the radial and angular position and velocity of the foot with respect to corresponidng hip. $T_C^P$ represents the transformation from the Cartesian space to the polar space.

\subsection{Highly structured policy}
While the desired foot positions and velocities of the structured policy
are directly mapped from robot states using a neural network policy, the highly structured policy exploits a
parameterized trajectory planner to design a trajectory (\autoref{fig:overview}c). As shown in \autoref{fig:controller}, the desired trajectory is parameterized by $T_\text{st}, T_\text{sw}, \bm{C}, l_\text{span}, \delta,$ and $\bm{P}_0$. 
Details of the shape and timing of the trajectory can be found in~\cite{lee_dynamics_2014, hyun_high_2014, hyun_implementation_2016}, in which all parameters were tuned manually. 

The trajectory planner composes of a gait pattern modulator which governs the temporal pattern of four legs based on a touchdown event of a reference leg (front left), and a leg trajectory generator which shapes the trajectory with a B\'ezier curve and a sinusoidal curve. 

The highly structured policy tries to learn the parameters of the gait modulator and trajectory generator that can scale a nominal trajectory based on the input desired velocity ($v_d$).

\begin{equation}
\begin{split}
    &\bm{a}^h_t = \pi^h( \bm{s}^h_t),\\
    &[k_T, \bm{k}_C, k_\delta, k_r, b_r, k_\theta, b_\theta] = \bm{a}^h_t,\\
    &\bm{s}^h_t = v_d
\end{split}
\end{equation}
where the desired velocity $v_d$ is the only input of the actor of the highly structured policy $\pi^h$, and the output action $\bm{a}^h_t$ is a series of control parameters. 

\begin{equation}
\begin{split}
    &[\bm{r}^\text{des}_{t}, \theta^\text{des}_{t}, \dot{r}^\text{des}_{t}, \dot{\theta}^\text{des}_{t}] = \\ &F(k_TT_\text{st}, k_TT_\text{sw}, \bm{k}_C\bm{C}, l_\text{span}, k_\delta\delta, \bm{P}_0),
\end{split}
\end{equation}

where $F$ denotes the trajectory planner. $k_T$ is the scaled ratio of $T_\text{st}, T_\text{sw}$; $\bm{k_C}$ denotes a scaled ratio vector for control points of the swing trajectory $\bm{C}$: $\bm{c}_0,\dots, \bm{c}_{11}$; $k_\delta$ is the scaled ratio of $\delta$. The meanings of all parameters are described in \autoref{fig:controller} and \autoref{tab:parameters}. It is not necessary to learn $l_\text{span}$ and $\bm{P}_0$ because they are constrained by other parameters (see \autoref{fig:controller}).

The low-level impedance controller of the highly structured policy is the same as that of the structured policy \eqref{eq:polar_impedance}.

\section{Learning Algorithms}
\label{sec:learning}
Both the direct policy and the structured policy were trained by a state-of-art reinforcement learning algorithm, twin delayed deep deterministic policy gradient algorithm (TD3)~\cite{fujimoto_addressing_2018}, which is based on the temporal difference (TD) learning~\cite{sutton_learning_1988}. The highly structured policy was trained by an one-step deep deterministic policy gradient (One-step DDPG) method rather than the TD method \cite{lillicrap_continuous_2015}. 

\subsection{TD3}
\label{subsec:TD3}
RL is based on a Markov decision process, which includes states $\bm{s}_t \in \mathcal{S}$, actions $\bm{a}_t \in \mathcal{A}$, stochastic transitions $p(\bm{s}_{t+1}|\bm{s}_t, \bm{a}_t)$, a reward function $r(\bm{s}_t, \bm{a}_t)$, and a discount factor $\gamma$.
The fundamental idea of RL is to train a policy $\bm{a}_t = \pi(\bm{s}_t): \mathcal{S}\rightarrow\mathcal{A}$ \ that selects the optimal action $\bm{a}_t$ based on the current states $\bm{s}_t$. The policy receives a reward for the state-action pair $r(\bm{s}_t, \bm{a}_t)$ and its objective is to maximize the expected cumulative reward $J$,

\begin{equation}\label{eq:expected_return}
    J = \mathbb{E}_{\bm{s}_{t+1} \thicksim p_{\pi}(\bm{s}_t), \bm{a}_t \thicksim \pi} \left[\sum_{t=0}^{T \text{--} 1} \gamma^t r_{t}(\bm{s}_t, \bm{a}_t) \right].
\end{equation}

The cumulative reward $J$ is denoted by a Q value in the standard Q-learning and is trained through the TD learning method \cite{sutton_learning_1988}, which is based on the Bellman equation:

\begin{equation}
\label{eq:bellman}
\begin{split}
    &Q_{\theta}(\bm{s}_t, \bm{a}_t) = r_{t} + \gamma \mathbb{E}_{\bm{s}_{t+1}, \bm{a}_{t+1}}[Q_{\theta}(\bm{s}_{t+1}, \bm{a}_{t+1})],\\
    &\bm{a}_{t+1} = \pi_\phi(\bm{s}_{t+1}),
\end{split}
\end{equation}
where $Q_{\theta}$ and $\pi_\phi$ indicate a critic with parameters $\theta$ and an actor with parameters $\phi$.

The standard actor-critic method based on the TD learning \eqref{eq:bellman} may overestimate the Q value because the maximum Q value at the next time step is adopted to update the current Q value. 
TD3, whose detailed implementation is shown in \cite{fujimoto_addressing_2018}, decreases the error of estimating the Q value by proposing two independent critics $Q_{\theta_1}$ and $Q_{\theta_2}$ and two corresponding target critic networks $Q_{\theta'_1}$ and $Q_{\theta'_2}$. The minimum of two target Q values is utilized to calculate the target $y$:

\begin{equation}
\label{eq:target_value}
\begin{split}
    &y = r_{t} + \gamma \min_{i=1, 2} Q_{\theta'_i}(\bm{s}_{t+1}, \bm{a}_{t+1}),\\
    &\bm{a}_{t+1} = \pi_{\phi}(\bm{s}_{t+1}) = \argmax_{\bm{a}_{t+1}} Q_{\theta'_1}(\bm{s}_{t+1}, \bm{a}_{t+1}).
\end{split}
\end{equation}

The mean squared error between the target value $y$ and the current Q values $Q_{\theta_i}(\bm{s}_t, \bm{a}_t)$ is set as the loss of the critics to track the target $y$:
\begin{equation}
\label{eq:loss}
\begin{split}
    \mathcal{L}(\theta_i) = \mathbb{E}_{\pi_{\phi}}[(Q_{\theta_i}(\bm{s}_t, \bm{a}_t)- y)^2].
\end{split}
\end{equation}

\subsection{\textcolor{black}{One-step DDPG}}
\label{subsec:ssac}
In this paper, an one-step learning method is utilized to directly estimate and maximize the cumulative reward in an episode. RL usually adopts TD learning rather than one-step learning because TD learning updates the estimation at each time step before the final outcome is known \cite{sutton_reinforcement_2018}. 

Nevertheless, TD learning is not necessary for the proposed highly structured policy. For TD learning, which is shown in \eqref{eq:bellman}, the current state-action pair ($\bm{s}_t, \bm{a}_t$) is different from the state-action pair at the next time step ($\bm{s}_{t+1}, \bm{a}_{t+1}$). Therefore, TD learning can adjust the estimation values for different action-state pairs at each time step, which increases the sampling efficiency. However, the proposed highly structured policy only receives the desired velocity $v_d$, which is constant in an episode. The output control parameters $\bm{a}$ only depend on the desired velocity $v_d$ and remain the same if the desired velocity $v_d$ does not change. There is only one state-action pair in an episode, and the cumulative reward will be known when the episode is finished. If the cumulative reward is known, it is not necessary to use the TD learning to update the estimation recursively. Using the one-step learning method, the critic network estimating the cumulative reward can be trained stably by supervised learning with the collected samples of actual cumulative reward. The detailed training process of the highly structured policy is shown in Algorithm \ref{alg:DDPG}.

\section{Simulation setup}
We compared the performance of direct, structured, and highly structured policies in learning a quadruped robot to track desired velocities. A planar, 11 degrees-of-freedom\footnote{Eight for each leg hip and knee joints, two for the center of mass horizontal and vertical position, and pitch orientation of a base platform.} robot model was built in MuJoCo~\cite{todorov_mujoco_2012}. The parameters of the model are listed in an
Appendix. We used the built-in integrator of MuJoCo, semi-implicit Euler, with simulation time step $\Delta t$ = 0.01 s.

The reward function at each time step was designed as
\begin{equation}
    r_t = 1-\frac{|v_t-v_d|}{v_d},
\end{equation}
where $v_t$ and $v_d$ are the center-of-mass velocity of the robot at each time step and the desired velocity, respectively.

\begin{algorithm}[H]
\caption{One-step DDPG}
\label{alg:DDPG}
\begin{algorithmic}
   \STATE Initialize the critic $Q_{\theta}$ and the actor $\pi_\phi$ using random parameters $\theta$ and $\phi$, and initialize replay buffer $\mathcal{B}$.
   \FOR{$t=1$ {\bfseries to} $T$}
   \IF{$t < T_\text{random}$}
   \STATE Generate parameters randomly $\bm{a} \thicksim \mathcal{U}(\bm{a}_{\min}, \bm{a}_{\max})$.
   \ELSE
   \STATE Generate parameters with exploration noise $\bm{a} \thicksim \pi_\phi(v_d) + \bm{\varepsilon}$, $\bm{\varepsilon} \thicksim \mathcal{N}(0, \sigma)$.
   \ENDIF
   \IF{an episode is not finished}
   \STATE Collect the reward: $R=\sum_{i=0}^{i_{\max}}r_i(v_d, \bm{a})$.
   \ELSE
   \STATE Store the state-action tuple $(v_d, \bm{a}, R)$ in $\mathcal{B}$. 
   \STATE Sample $N$ tuples $(v_d, \bm{a}, R)$ from $\mathcal{B}$.
   \FOR{$k=1$ {\bfseries to} $K$}
   \STATE Update critic $\theta \leftarrow \arg \min_{\theta} N^{ \text{--} 1} \sum(R - Q_{\theta}(v_d,\bm{a}))^2$.
   \ENDFOR
   \FOR{$k=1$ {\bfseries to} $K$}
   \STATE Update the actor $\pi_\phi$ by the deterministic policy gradient $\nabla_{\phi} J(\phi)$:
   \STATE $N^{ \text{--} 1} \sum [-\nabla_{\bm{a}} Q_{\theta}(v_d,\bm{a}) |_{\bm{a}=\pi_{\phi}(v_d)} \nabla_{\phi} \pi_\phi(v_d)]$.
   \ENDFOR
   \STATE Change the desired velocity $v_d \thicksim \mathcal{U}(v_{\min}, v_{\max})$.
   \ENDIF
   \ENDFOR
\end{algorithmic}
\end{algorithm}

We also implemented an `early stop' condition, with which the simulator detected a failure of locomotion and terminated each episode during training if excessive pitch rotation or the center-of-mass height decrease was detected ($|\text{pitch}_t|>1 $ rad or $h^\text{com}_t < 0.3$ m).  This condition changes the cost function to discourage undesirable behaviors  \cite{peng_deepmimic:_2018}.

All policies were trained for $3\times10^5$ time steps and evaluated every 5000 time steps. The maximum time steps in each episode were 1000, corresponding to 10 seconds in the simulation. If the episode does not meet the early stop condition, the number of total iterations is 300 ($3\times10^5/1000$). The desired velocity changed randomly from 1 m/s to 5 m/s at the start of each episode in both the training and evaluation stage. In the evaluation stage, 10 episodes were repeated without exploration noise to calculate the mean of the cumulative rewards. The policy with the highest mean of the cumulative reward was saved as the best policy. Each policy was trained five times with different random seeds. The hyperparameters of training different policies are listed in an
Appendix. A computer with an Intel i7-9700K CPU and an NVIDIA GeForce RTX2070 GPU was used to train policies.

Once the best parameters for each policy were determined, we tested the performance of each policy in the task of tracking different velocities under disturbances. Each simulation was run with an initial desired velocity of 2 m/s, changed to 4 m/s at 4 s, then to 3 m/s at 8 s. A pushing (forward) force with 10 N and a pulling (backward) force with 10 N were applied to the center of mass of the model to test the robustness of each policy, at 12 s and 16 s, respectively.

\section{Simulation Results}
\subsection{Training result}
\begin{figure}[htpb]
    \centering
    \includegraphics[width=\columnwidth]{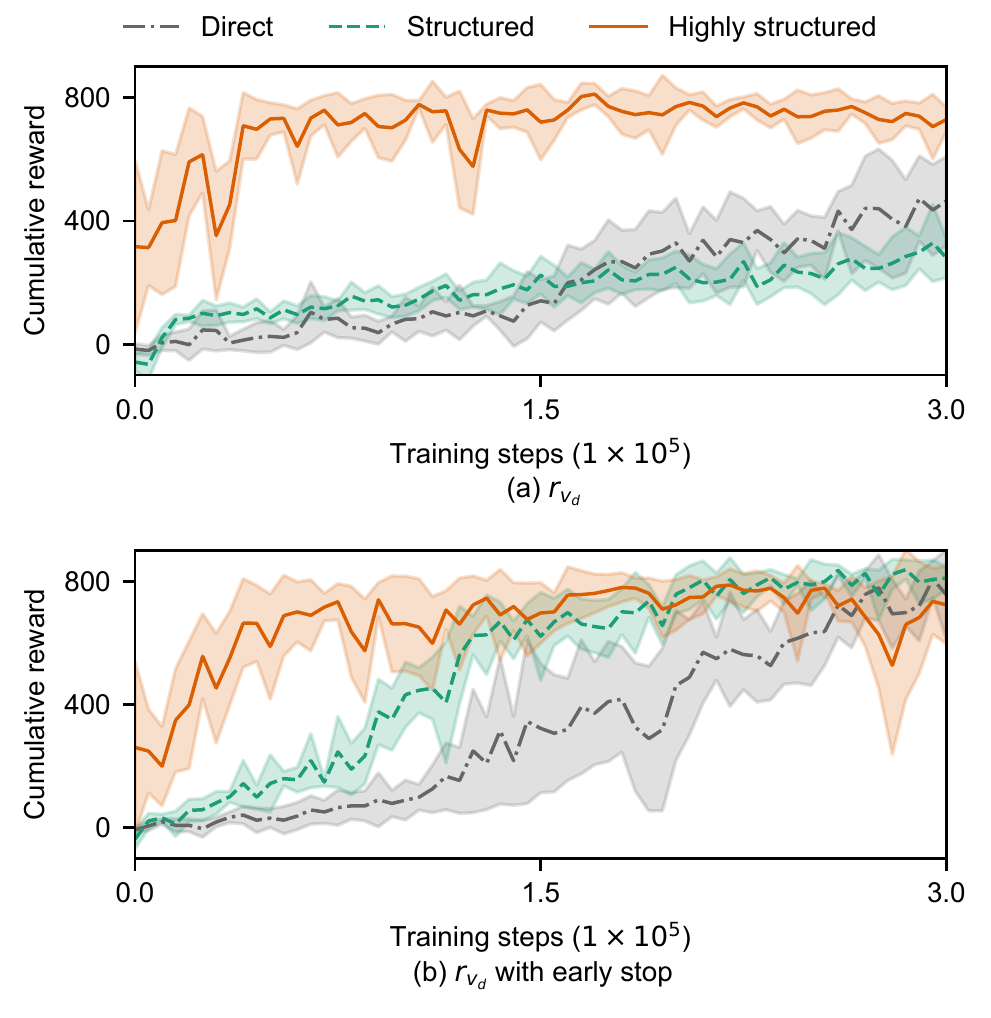}
    \caption{Received cumulative reward in the training process. The line and the shaded region represent the mean and standard deviation of the average evaluation in five trials. Two figures show the results of different cost functions.} 
    \label{fig:reward}
\end{figure}

The results of trained policies are summarized in \autoref{tab:performances}.
The learning curves of each policy are shown in \autoref{fig:reward}. The results show that the highly structured policy is not sensitive to different cost functions. Conversely, both the
direct and structured policies show different performance for different cost functions. 
If a cost function (e.g., $r_{v_d}$ with early stop) was
chosen appropriately, the structured policy could achieve a higher cumulative reward than the highly structured policy after exploring a long time (\autoref{fig:reward}b and \autoref{tab:performances}). 
\textcolor{black}{
In the following sections, we only compare the trained policies using the cost function with early stop because all three policies converged, which is shown in \autoref{fig:reward}b.
}

\textcolor{black}{
The highly structured policy converged much more quickly than the other two policies, and the structured policy converged more quickly than the direct policy if the cost function was suitable (\autoref{fig:reward}b). The convergence rate is represented by the rise steps and rise time in \autoref{tab:performances}.
}
The rise steps and rise time denote the required simulation steps and computing time to train a policy to receive an average cumulative reward higher than 700, which is 70\% of the maximum possible cumulative reward.

\begin{table}[ht]
\centering
\caption {\label{tab:performances} Performance of different policies}
\renewcommand{\arraystretch}{1} 
\begin{center}
\begin{tabular}{l c c c}
\toprule
Policy & Direct & Structured & Highly structured \\
\midrule
Rise steps ($\times10^5$) & 2.6 & 1.8 & \textbf{0.6} \\
Rise time (minutes) & 65 & 45 & \textbf{3} \\
Highest reward & 821.17 & \textbf{867.53} & 832.08 \\
Std of pitch (rad) & 0.86 & 0.17 & \textbf{0.09} \\
Std of height (m) & 0.14 & 0.08 & \textbf{0.04}\\
COT & 4.97 & 3.98 & \textbf{2.63} \\
\end{tabular}
\end{center}
\end{table}

\subsection{Velocity tracking under force disturbance}
\label{subsec:velocity}
\begin{figure}[!h]
    \centering
    \includegraphics[width=\columnwidth]{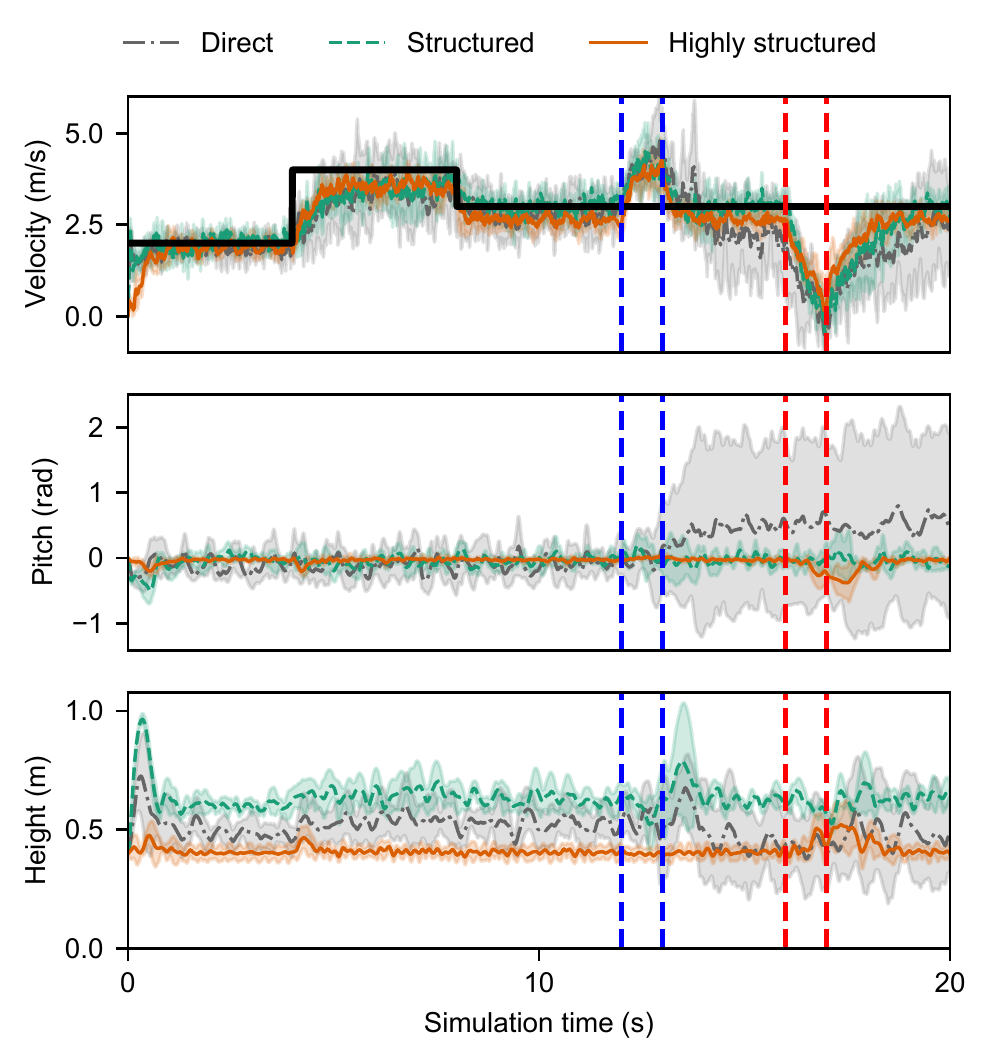}
    \caption{Tracking different velocities under disturbances. The line and the shaed areas indicate the mean and a standard deviation of the values in five trials. The black solid line indicates the desired velocity $v_d$. The blue dash line and the red dash line represent a 10 N forward disturbance force and a 10 N backward disturbance force, respectively.} 
    \label{fig:velocity}
\end{figure}

\autoref{fig:velocity} shows that both the structured and the highly structured policy could track different desired velocities and recover after force disturbances were applied. On the other hand, the direct policy became unstable under the disturbances. 

The variation of body height and pitch are indicative of locomotion stability. As shown in \autoref{fig:velocity} and \autoref{tab:performances}, the standard deviation (std) of the body height and pitch of the robot trained with the highly structured policy were lower than that with the other two policies. In particular, the robot sometimes tumbled ($|\text{pitch}_t| > 1 \text{ rad} \text{ or } |h^\text{com}_t| < 0.3 \text{ m}$) under disturbances with the direct policy. 

\subsection{Energy efficiency}
While the energetic efficiency was not the objective of training, the cost of transport was compared to assess the quality of the resultant behaviors. The COT was calculated as $\text{COT}=\frac{\bar{P}}{W\bar{v}}$, where $\bar{P}$ is the average power, $W$ is the weight of the robot, and $\bar{v}$ is the average velocity \cite{hyun_high_2014}. As shown in \autoref{tab:performances}, the more structured the policy was, the more efficient the resultant behavior was.

\subsection{Gait periodicity}
\begin{figure}[htpb]
    \centering
    \includegraphics[width=\columnwidth]{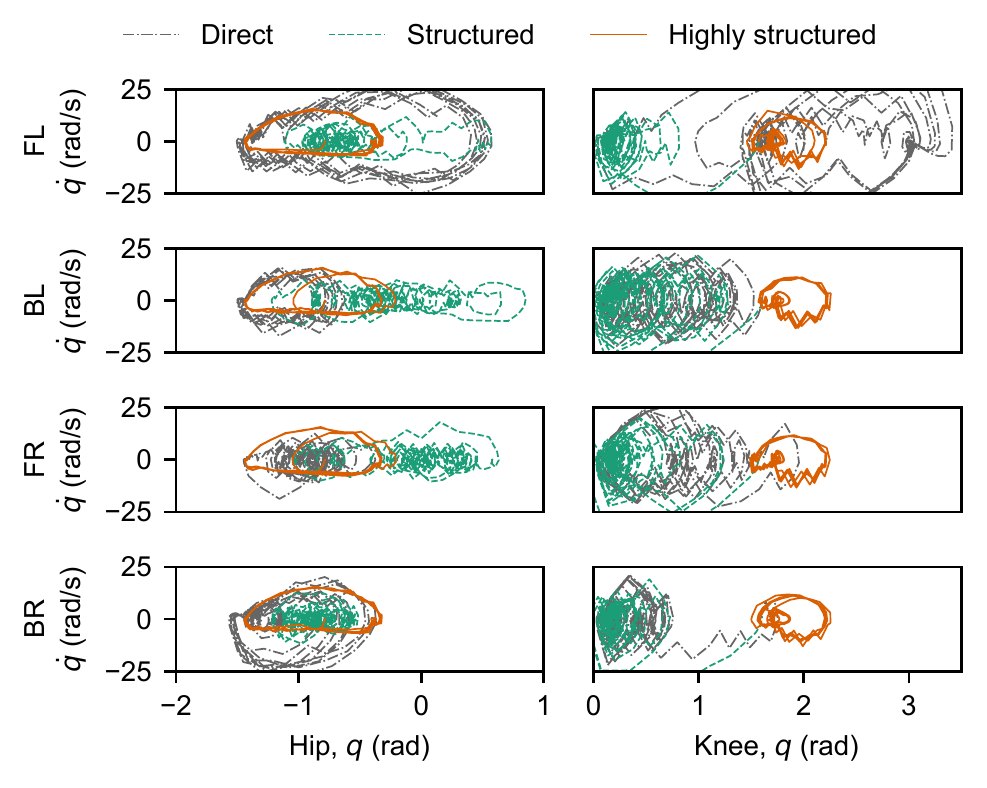}
    \caption{Joint periodic orbits of learned policies for tracking a constant velocity: 2 m/s.} 
    \label{fig:limit_cycle}
\end{figure}

\autoref{fig:limit_cycle} shows the trajectory of all four legs with different policies while the robot was tracking a constant desired velocity (2 m/s). As shown in \autoref{fig:limit_cycle}, the highly structured policy had a repetitive pattern, while the other two policies showed highly variable trajectories; any
pattern, if it existed,
was not clear.

\section{Discussion}
\textcolor{black}{
The present study incorporated different structures with the direct policy to improve its performance on learning quadruped locomotion and investigated the quantitative effect of embedded structures on the performance of the policy. 
} 
Simulation results validated that the highly structured policy could decrease the required training time and sensitivity to the cost function,
and achieve more robust and energetically efficient gaits. There were subtle but important differences in the performance of direct and structured policies as well. 


Our results replicated
previous findings that different action spaces resulted in different performance for learning robotic tasks~\cite{varin_comparison_2019, bogdanovic_learning_2019}.  
In principle, the direct policy could achieve better performance compared to its structured alternative because it has no constraint. In practice, from the perspective of optimization, finding the global optimum is rarely possible especially if the objective function is complex. To the best knowledge of the authors, legged locomotion is a complicated task as the objective function usually becomes highly nonlinear, non-convex, non-smooth, and even discontinuous due to intermittent contact between the leg and the ground. Hence it seems that
using the direct policy for learning legged locomotion has very little advantage. Employing a structured controller that can effectively limit the
search space and formulate the problem in a feasible way is
beneficial. 

The highly structured policy outperformed the other two policies. What was the essential difference between the structured and the highly structured policy presented in this work? First of all, unlike the direct and the structured policy, whose actor network has hundreds of thousands of trainable parameters, the complicated structure of the last approach reduced the dimension of parameters to train to 32. Moreover, as stated in \autoref{sec:learning}, the direct and the structured policies were trained by the TD3, which required to train the network at each time step, while the highly structured policy only requires to train the network after finishing each episode. Hence, the rise time for the highly structured policy could be decreased to 3 minutes, which was about 5\% of that required for
the direct policy. 

We posit that a more important difference is the explicit representation of the planner in the highly structured policy, which provides proper timing to the impedance controller. There is substantial
evidence that animal legged locomotion is generated and controlled by feedback mechanisms interacting with a central pattern generator (CPG)~\cite{delcomyn1980neural,grillner1985central}. This 
network of neurons is intrinsically capable of producing rhythmic output without requiring sensory feedback. The CPG, processing forward-path information, and peripheral neuro-mechanics, coping with interactive dynamics,
have fundamentally distinctive structure~\cite{hogan2017physical}, and both may be equally important for successful locomotion. Our task-space impedance controller copes with interactive dynamics. In addition, the task-space planner in the highly structured policy may correspond to the CPG. Lacking a forward-path rhythmic signal generator in the structured policy may have degraded its performance.

While implementing the
early stop condition is a common technique used in RL for robotics to reduce total training time~\cite{fujimoto_addressing_2018, peng_deepmimic:_2018},  the highly structured control was largely unaffected by this condition. Moreover, in our preliminary experiments, if either excessive pitch or excessive height was removed from the failure detection, the direct and the structured policy did not perform well. This seems attributable to the complex mechanics of legged locomotion; the system is floating-based and under-actuated. In summary, the robot may lose its controllability in certain configurations, for example when all four legs are off the ground. As the direct and the structured policy do not incorporate this complexity in learning locomotion, it is necesssary to provide additional information in the form of an
early stop condition to prevent policies from exploring infeasible regions in state-action spaces. On the other hand, the highly structured control was already designed to avoid such failures, and thus the early stop condition was not important~\cite{hyun_high_2014}. 
Likewise, the energetically efficient and periodic gait pattern emerged from the highly structured policy as properties of its sophisticated controller.

Are we concluding that the highly structured policy with more details is always the best? Not exactly. \textcolor{black}{
The performance, including the robustness and energy cost, of the highly structured policy heavily depends on the control structure. If a different control structure is utilized, the highly structured policy may have notably different properties. Therefore, a proper control structure should be carefully chosen before designing the highly structured policy.
}
The highly structured policy we used in the present study is tailored to a specific task: trotting of a planar quadruped robot. It cannot be generalized to a broader class of robots (hexapods with more legs or bipeds with fewer legs, etc.), different forms of running (gallop, pronk, bound, etc.), nor it is not straightforward how to expand this controller to 3D robots. We speculate that there exists a sweet-spot between these two extremes - the direct and the highly structured. 

What we have discussed so far are different ways of embedding structure in the machine learning problem. We may design a sophisticated controller tailored to desirable behavior. Instead, we may embed structure by properly designing cost functions and hard-coding constraints in state-action spaces. Combining these two by identifying and extracting proper state-action spaces from successful controllers and designing policy appropriately might be the sweet spot. In pursuing this, a favorable approach seems to incorporate insight and intuition from previous engineering experience that are abundant in the literature. 

\section{Conclusion}
\textcolor{black}{
In the present paper, different structures were incorporated to improve the performance of the policy on the task of learning quadruped locomotion to track desired velocities, and the quantitative effects of different structures were analyzed. The simulation results showed that the direct and the structured policy required additional consideration on balance in designing reward for training algorithm while highly structured policy did not. Both the structured policy and the highly structured policy showed better performance in terms of sampling efficiency, velocity error, robustness, and energy efficiency than the direct policy. In particular, the structured policy and the highly structured policy required 1/3  and 3/4 fewer simulation steps to achieve a high cumulative reward, compared to the direct policy. The high performance of the highly structured policy should be attributable to the reduced state-action space of the policy by incorporating engineering insights for successful quadruped locomotion in the structured policy. It seems beneficial to inherit physical intuitions and insights underlying previous controllers and adapt its structure to simplify the learning problems rather than using direct policies for these highly complicated robotic tasks. 
}

\begin{appendix}
\label{sec:appendix}
\subsection{Robot model parameters}
The total mass of the robot is 14 kg. Each segment mass was determined based on its volume. The size and mechanical joint limits are listed in \autoref{tab:robot_model}.

\begin{table}[!h]
\centering
\caption {\label{tab:robot_model} Model parameters}
\renewcommand{\arraystretch}{1} 
\begin{center}
\begin{tabular}{l l l}
\toprule
Segment & Length (m) & Radius (m)\\
\midrule
Body & 0.6 & 0.05\\
Thigh & 0.283 & 0.03\\
Shank & 0.283 & 0.03\\
\toprule
Joint & Motion range ($^\circ$) & Maximum torque (N$\cdot$m)\\
\midrule
Hip & [-80, 80] & 100\\
Knee & [10, 170] & 100\\
\end{tabular}
\end{center}
\end{table}

\subsection{Training hyperparameters}
The hyperparameters used to train the TD3 and one-step DDPG are shown in \autoref{tab:training_parameters}.

\begin{table}[!h]
\centering
\caption {\label{tab:training_parameters} Training hyperparameters}
\renewcommand{\arraystretch}{1} 
\begin{center}
\resizebox{1\columnwidth}{!}{
\begin{tabular}{l c c}
\toprule
Algorithm & TD3\cite{fujimoto_addressing_2018} & One-step DDPG \cite{lillicrap_continuous_2015}\\
\midrule
Optimizer & Adam & Adam\\
Learning rate & 0.001 & 0.001\\
Discount $\gamma$ & 0.99 & /\\
Exploration noise & 0.1 & 0.1\\
Total time steps & $3\times10^5$ & $3\times10^5$\\
Buffer size & $3\times10^5$ & $3\times10^5$\\
Batch size & 100 & 100\\
Training steps in each episode & / & 100\\
Time steps of random exploration & $1\times10^4$ & $1\times10^4$\\
Target network update rate & 0.005 & /\\
Target policy noise & 0.2 & /\\
Delayed time steps of policy updates & 2 & /\\
\end{tabular}
}
\end{center}
\end{table}
\end{appendix}

\bibliographystyle{IEEEtran}
\bibliography{main}

\begin{thebibliography}{10}
\providecommand{\url}[1]{#1}
\csname url@samestyle\endcsname
\providecommand{\newblock}{\relax}
\providecommand{\bibinfo}[2]{#2}
\providecommand{\BIBentrySTDinterwordspacing}{\spaceskip=0pt\relax}
\providecommand{\BIBentryALTinterwordstretchfactor}{4}
\providecommand{\BIBentryALTinterwordspacing}{\spaceskip=\fontdimen2\font plus
\BIBentryALTinterwordstretchfactor\fontdimen3\font minus
  \fontdimen4\font\relax}
\providecommand{\BIBforeignlanguage}[2]{{%
\expandafter\ifx\csname l@#1\endcsname\relax
\typeout{** WARNING: IEEEtran.bst: No hyphenation pattern has been}%
\typeout{** loaded for the language `#1'. Using the pattern for}%
\typeout{** the default language instead.}%
\else
\language=\csname l@#1\endcsname
\fi
#2}}
\providecommand{\BIBdecl}{\relax}
\BIBdecl

\bibitem{hyun_high_2014}
D.~J. Hyun, S.~Seok, J.~Lee, and S.~Kim, ``\BIBforeignlanguage{en}{High speed
  trot-running: implementation of a hierarchical controller using
  proprioceptive impedance control on the mit cheetah},''
  \emph{\BIBforeignlanguage{en}{The International Journal of Robotics
  Research}}, vol.~33, no.~11, pp. 1417--1445, Sep. 2014.

\bibitem{fu_gait_2008}
C.~Fu and K.~Chen, ``Gait synthesis and sensory control of stair climbing for a
  humanoid robot,'' \emph{IEEE Transactions on Industrial Electronics},
  vol.~55, no.~5, pp. 2111--2120, May 2008.

\bibitem{neunert_whole-body_2018}
M.~Neunert, M.~Stäuble, M.~Giftthaler, C.~D. Bellicoso, J.~Carius, C.~Gehring,
  M.~Hutter, and J.~Buchli, ``Whole-body nonlinear model predictive control
  through contacts for quadrupeds,'' \emph{IEEE Robotics and Automation
  Letters}, vol.~3, no.~3, pp. 1458--1465, Jul. 2018.

\bibitem{magana_fast_2019}
O.~A.~V. Magaña, V.~Barasuol, M.~Camurri, L.~Franceschi, M.~Focchi, M.~Pontil,
  D.~G. Caldwell, and C.~Semini, ``Fast and continuous foothold adaptation for
  dynamic locomotion through cnns,'' \emph{IEEE Robotics and Automation
  Letters}, vol.~4, no.~2, pp. 2140--2147, Apr. 2019.

\bibitem{fahmi_passive_2019}
S.~Fahmi, C.~Mastalli, M.~Focchi, and C.~Semini, ``Passive whole-body control
  for quadruped robots: experimental validation over challenging terrain,''
  \emph{IEEE Robotics and Automation Letters}, vol.~4, no.~3, pp. 2553--2560,
  Jul. 2019.

\bibitem{lillicrap_continuous_2015}
T.~P. Lillicrap, J.~J. Hunt, A.~Pritzel, N.~Heess, T.~Erez, Y.~Tassa,
  D.~Silver, and D.~Wierstra, ``Continuous control with deep reinforcement
  learning,'' \emph{arXiv:1509.02971 [cs, stat]}, Sep. 2015.

\bibitem{xu_feedback_2019}
J.~Xu, Z.~Hou, W.~Wang, B.~Xu, K.~Zhang, and K.~Chen, ``Feedback deep
  deterministic policy gradient with fuzzy reward for robotic multiple
  peg-in-hole assembly tasks,'' \emph{IEEE Transactions on Industrial
  Informatics}, vol.~15, no.~3, pp. 1658--1667, Mar. 2019.

\bibitem{chen_deep_2019}
J.~Chen, T.~Shu, T.~Li, and C.~W. de~Silva, ``Deep reinforced learning tree for
  spatiotemporal monitoring with mobile robotic wireless sensor networks,''
  \emph{IEEE Transactions on Systems, Man, and Cybernetics: Systems}, pp.
  1--15, 2019.

\bibitem{zhang_teach_2019}
K.~Zhang, Z.~Hou, C.~W. de~Silva, H.~Yu, and C.~Fu, ``Teach biped robots to
  walk via gait principles and reinforcement learning with adversarial
  critics,'' \emph{arXiv:1910.10194 [cs]}, Oct. 2019.

\bibitem{peng_deeploco_2017}
X.~B. Peng, G.~Berseth, K.~Yin, and M.~Van De~Panne, ``{DeepLoco}: dynamic
  locomotion skills using hierarchical deep reinforcement learning,'' \emph{ACM
  Transactions on Graphics (TOG)}, vol.~36, no.~4, pp. 41:1--41:13, Jul. 2017.

\bibitem{peng_deepmimic:_2018}
X.~B. Peng, P.~Abbeel, S.~Levine, and M.~van~de Panne,
  ``\BIBforeignlanguage{en}{{DeepMimic}: example-guided deep reinforcement
  learning of physics-based character skills},''
  \emph{\BIBforeignlanguage{en}{ACM Transactions on Graphics}}, vol.~37, no.~4,
  pp. 1--14, Jul. 2018.

\bibitem{xie_iterative_2019}
Z.~Xie, P.~Clary, J.~Dao, P.~Morais, J.~Hurst, and M.~van~de Panne, ``Iterative
  reinforcement learning based design of dynamic locomotion skills for
  cassie,'' \emph{arXiv:1903.09537 [cs]}, Mar. 2019.

\bibitem{hwangbo_learning_2019}
J.~Hwangbo, J.~Lee, A.~Dosovitskiy, D.~Bellicoso, V.~Tsounis, V.~Koltun, and
  M.~Hutter, ``\BIBforeignlanguage{en}{Learning agile and dynamic motor skills
  for legged robots},'' \emph{\BIBforeignlanguage{en}{Science Robotics}},
  vol.~4, no.~26, p. eaau5872, Jan. 2019.

\bibitem{singla_realizing_2019}
A.~Singla, S.~Bhattacharya, D.~Dholakiya, S.~Bhatnagar, A.~Ghosal, B.~Amrutur,
  and S.~Kolathaya, ``Realizing learned quadruped locomotion behaviors through
  kinematic motion primitives,'' in \emph{2019 {International} {Conference} on
  {Robotics} and {Automation} ({ICRA})}, May 2019, pp. 7434--7440.

\bibitem{haarnoja_soft_2018}
T.~Haarnoja, A.~Zhou, P.~Abbeel, and S.~Levine, ``\BIBforeignlanguage{en}{Soft
  actor-critic: off-policy maximum entropy deep reinforcement learning with a
  stochastic actor},'' in \emph{\BIBforeignlanguage{en}{International
  {Conference} on {Machine} {Learning}}}, Jul. 2018, pp. 1861--1870.

\bibitem{fujimoto_addressing_2018}
S.~Fujimoto, H.~Hoof, and D.~Meger, ``\BIBforeignlanguage{en}{Addressing
  function approximation error in actor-critic methods},'' in
  \emph{\BIBforeignlanguage{en}{International {Conference} on {Machine}
  {Learning}}}, Jul. 2018, pp. 1587--1596.

\bibitem{hou_off-policy_2020}
Z.~Hou, K.~Zhang, Y.~Wan, D.~Li, C.~Fu, and H.~Yu, ``Off-policy maximum entropy
  reinforcement learning: soft actor-critic with advantage weighted mixture
  policy(sac-awmp),'' \emph{arXiv:2002.02829 [cs, stat]}, Feb. 2020.

\bibitem{jain_hierarchical_2019}
D.~Jain, A.~Iscen, and K.~Caluwaerts, ``Hierarchical reinforcement learning for
  quadruped locomotion,'' \emph{arXiv:1905.08926 [cs]}, May 2019.

\bibitem{varin_comparison_2019}
P.~Varin, L.~Grossman, and S.~Kuindersma, ``A comparison of action spaces for
  learning manipulation tasks,'' in \emph{2019 {IEEE}/{RSJ} {International}
  {Conference} on {Intelligent} {Robots} and {Systems} ({IROS})}, Nov. 2019,
  pp. 6015--6021.

\bibitem{tsounis_deepgait_2020}
V.~Tsounis, M.~Alge, J.~Lee, F.~Farshidian, and M.~Hutter, ``Deepgait: planning
  and control of quadrupedal gaits using deep reinforcement learning,''
  \emph{IEEE Robotics and Automation Letters}, vol.~5, no.~2, pp. 3699--3706,
  Apr. 2020.

\bibitem{buchli_learning_2011}
J.~Buchli, F.~Stulp, E.~Theodorou, and S.~Schaal,
  ``\BIBforeignlanguage{en}{Learning variable impedance control},''
  \emph{\BIBforeignlanguage{en}{The International Journal of Robotics
  Research}}, vol.~30, no.~7, pp. 820--833, Jun. 2011.

\bibitem{bogdanovic_learning_2019}
M.~Bogdanovic, M.~Khadiv, and L.~Righetti, ``Learning variable impedance
  control for contact sensitive tasks,'' \emph{arXiv:1907.07500 [cs]}, Jul.
  2019.

\bibitem{martin-martin_variable_2019}
R.~Martín-Martín, M.~A. Lee, R.~Gardner, S.~Savarese, J.~Bohg, and A.~Garg,
  ``Variable impedance control in end-effector space: an action space for
  reinforcement learning in contact-rich tasks,'' \emph{arXiv:1906.08880 [cs]},
  Aug. 2019.

\bibitem{lee_dynamics_2014}
J.~Lee, D.~J. Hyun, J.~Ahn, S.~Kim, and N.~Hogan, ``On the dynamics of a
  quadruped robot model with impedance control: self-stabilizing high speed
  trot-running and period-doubling bifurcations,'' in \emph{2014 {IEEE}/{RSJ}
  {International} {Conference} on {Intelligent} {Robots} and {Systems}}, Sep.
  2014, pp. 4907--4913.

\bibitem{hyun_implementation_2016}
D.~J. Hyun, J.~Lee, S.~Park, and S.~Kim,
  ``\BIBforeignlanguage{en}{Implementation of trot-to-gallop transition and
  subsequent gallop on the mit cheetah {I}},''
  \emph{\BIBforeignlanguage{en}{The International Journal of Robotics
  Research}}, vol.~35, no.~13, pp. 1627--1650, Nov. 2016.

\bibitem{sutton_learning_1988}
R.~S. Sutton, ``\BIBforeignlanguage{en}{Learning to predict by the methods of
  temporal differences},'' \emph{\BIBforeignlanguage{en}{Machine Learning}},
  vol.~3, no.~1, pp. 9--44, Aug. 1988.

\bibitem{sutton_reinforcement_2018}
R.~S. Sutton and A.~G. Barto, \emph{Reinforcement learning: {An}
  introduction}.\hskip 1em plus 0.5em minus 0.4em\relax MIT press, 2018.

\bibitem{todorov_mujoco_2012}
E.~Todorov, T.~Erez, and Y.~Tassa, ``{MuJoCo}: {A} physics engine for
  model-based control,'' in \emph{2012 {IEEE}/{RSJ} {International}
  {Conference} on {Intelligent} {Robots} and {Systems}}, Oct. 2012, pp.
  5026--5033.

\bibitem{delcomyn1980neural}
F.~Delcomyn, ``Neural basis of rhythmic behavior in animals,'' \emph{Science},
  vol. 210, no. 4469, pp. 492--498, 1980.

\bibitem{grillner1985central}
S.~Grillner and P.~Wallen, ``Central pattern generators for locomotion, with
  special reference to vertebrates,'' \emph{Annual review of neuroscience},
  vol.~8, no.~1, pp. 233--261, 1985.

\bibitem{hogan2017physical}
N.~Hogan, ``Physical interaction via dynamic primitives,'' in \emph{Geometric
  and Numerical Foundations of Movements}.\hskip 1em plus 0.5em minus
  0.4em\relax Springer, 2017, pp. 269--299.

\end{thebibliography}
\end{document}